\documentclass[runningheads]{llncs}
\usepackage{amsmath}
\usepackage[T1]{fontenc}
\usepackage{amssymb}
\usepackage{graphicx}
\usepackage{listings}
\usepackage{url}
\usepackage{colortbl} 
\usepackage{array}     
\usepackage{xcolor}   
\usepackage{booktabs} 
\usepackage{marvosym}

\begin{document}

\mainmatter
\title{Vision Language Models for Dynamic Human Activity Recognition in Healthcare Settings}


\author{Abderrazek Abid\inst{1} \and Thanh-Cong Ho\inst{1}\textsuperscript{(\Letter)} \and Fakhri Karray\inst{1,2}\textsuperscript{(\Letter)}}

\authorrunning{Abid et al.}
\titlerunning{Dynamic Human Activity Recognition in Healthcare Settings}

\institute{
MBZUAI, Abu Dhabi, UAE \\
\email{\{abid.abderrazek, thanh.ho, fakhri.karray\}@mbzuai.ac.ae}
\and
University of Waterloo, Waterloo, ON, Canada \\
\email{karray@uwaterloo.ca}
}
\maketitle

\begin{abstract}

As generative AI continues to evolve, Vision Language Models (VLMs) have emerged as promising tools in various healthcare applications. One area that remains relatively underexplored is their use in human activity recognition (HAR) for remote health monitoring. VLMs offer notable strengths, including greater flexibility and the ability to overcome some of the constraints of traditional deep learning models. However, a key challenge in applying VLMs to HAR lies in the difficulty of evaluating their dynamic and often non-deterministic outputs. To address this gap, we introduce a descriptive caption data set and propose comprehensive evaluation methods to evaluate VLMs in HAR. Through comparative experiments with state-of-the-art deep learning models, our findings demonstrate that VLMs achieve comparable performance and, in some cases, even surpass conventional approaches in terms of accuracy. This work contributes a strong benchmark and opens new possibilities for the integration of VLMs into intelligent healthcare systems. Code and dataset are available at: \url{https://github.com/gouga10/VLMs-HAR-RHMS.git}


\end{abstract}

\section{Introduction}

Remote health monitoring has become an increasingly important area in healthcare, particularly in response to the growing elderly population. With the emergence of generative AI, there is an increasing expectation of intelligent systems that can continuously monitor patients while preserving their privacy. One promising approach is to encode visual data and employ AI models to infer patient activities, eliminating the need for direct access to raw images. In such systems, clinicians could query a vision-language model (VLM) with questions like “What is the patient doing?”, making human activity recognition (HAR) a vital component with significant potential to enhance healthcare delivery.

Despite the progress made by deep learning models in HAR~\cite{videoswin,pi,assemble}, several limitations remain. These models typically require extensive labeled datasets and are constrained by a fixed set of predefined activity classes. Furthermore, integrating separate models solely for HAR within a broader AI-assisted remote monitoring system, such as REMONI~\cite{Ho-2024}, introduces inefficiencies, especially when generative AI is already in use to facilitate interactions between doctors and the system. This creates a strong motivation to explore the use of VLMs for HAR, allowing the same model to both interpret patient activities and support natural-language interactions with healthcare providers.

Unlike traditional HAR models, Vision Language Models (VLMs) can generate more detailed and flexible descriptions of patient activities. Trained on large-scale multimodal datasets, VLMs can generalize across a broad range of actions without being constrained by predefined class labels. However, their adoption in healthcare remains limited, primarily due to the lack of standardized benchmarks for evaluating their recognition performance comprehensively and equitably.

Since VLMs generate free-form text, rigid keyword-based metrics may unfairly penalize them, while overly relaxed criteria risk inflating performance. To address this, our work introduces a caption-based dataset derived from the Toyota Smarthome video dataset~\cite{Das_2019_ICCV}, tailored to support visual-text alignment in healthcare monitoring scenarios. We also propose evaluation methods designed to fairly and comprehensively assess the ability of VLMs to recognize human activities.

\section{Related Work}

\subsection{Remote Health Monitoring Systems}

With the increasing burden on hospitals and advances in technology, numerous studies have been conducted on remote health monitoring systems (RHMS). The objective is to develop systems capable of monitoring the vital signs of multiple patients simultaneously, thereby alleviating the workload of medical professionals, enhancing patient care, and ensuring timely detection of health abnormalities to prevent missed emergency interventions.

Abu-Jassar et al. \cite{Abu-Jassar2024} proposed a system integrating hardware modules, sensors, and Amazon Web Services to track patient's vital signs. Meanwhile, Zhang et al. \cite{Zhang-2020} and Sufian et al. \cite{Sufian-2021} incorporated machine learning and deep learning models into RHMS to enhance system intelligence. Recently, with advancements in LLMs, Lee et al. \cite{Lee-2024} introduced a smart mirror that serves as a monitoring platform and provides interactive support for elderly individuals living alone. 

Similarly, Ho et al.~\cite{Ho-2024} introduced an IoT-enabled remote health monitoring system that incorporates a Natural Language Processing engine built on a Retrieval-Augmented Generation (RAG) framework to support clinicians in identifying health anomalies and accessing relevant patient information. Notably, their work was among the first to explore the application of VLMs for HAR within such systems. However, the evaluation methodology relied solely on the presence of ground-truth keywords in the generated output, which does not accurately reflect the full capabilities of VLMs. In addition, the study did not include a comparative analysis between VLMs and conventional deep learning models for activity recognition, leaving a gap in assessing their relative performance.

\subsection{Human Activity Recognition}
Activity recognition has been extensively studied for years, with numerous surveys exploring the evolution of methods, datasets, and models. Early works focused on convolutional neural networks (CNN) and long-short-term memory (LSTM) networks. In both CNN and transformer-based models, activity recognition has traditionally been framed as a classification task, where the model is constrained to choosing from a predefined and restricted set of activity labels. However, the introduction of VLMs \cite{VLM-HAR,chen2024internvlscalingvisionfoundation,llava} has transformed this paradigm by integrating visual and textual modalities, enabling open-ended activity descriptions rather than a finite set of labels. This allows VLMs to recognize and describe an unlimited range of activities, even those not explicitly seen during training, by leveraging the generative capabilities of language models and their contextual reasoning. However, since most of the datasets were created for conventional deep learning algorithms, labels usually consist only of class names, which is insufficient for VLMs. As a result, we found it necessary to create our dataset with descriptive textual captions for each visual sample

\section{Human Activity Descriptive Caption Dataset}

\subsection{Visual Datasets}
Although many visual datasets have been released for human activity recognition tasks, the Toyota Smarthome Dataset~\cite{Das_2019_ICCV} is selected for its close resemblance to real-world scenarios in RHMS.

The trimmed version of this dataset comprises 31 real-world activities that humans engage in daily, such as reading, watching TV, making coffee or breakfast, among others. Humans frequently interact with objects in this dataset (e.g., "drink from a can" and "cook cut"). The dataset includes 18 participants between the ages of 60 and 80, making it particularly relevant for RHMS, which are commonly designed to support elderly populations.

The dataset contains 16,115 videos across 31 action classes, captured from seven different camera viewpoints. For this study, only RGB frames are utilized, although depth and skeleton inputs are also available.

\subsection{Methodology for Descriptive Caption Generation}
VLMs are capable of describing human activities in greater detail compared with conventional deep learning models. Therefore, having textual ground truths would provide a more appropriate basis for evaluating the outputs of VLMs. In this section, descriptive captions corresponding to the ground truth labels for each video in the dataset are generated for evaluation purposes.

\begin{figure}
    \centering
    \includegraphics[width=1\columnwidth]{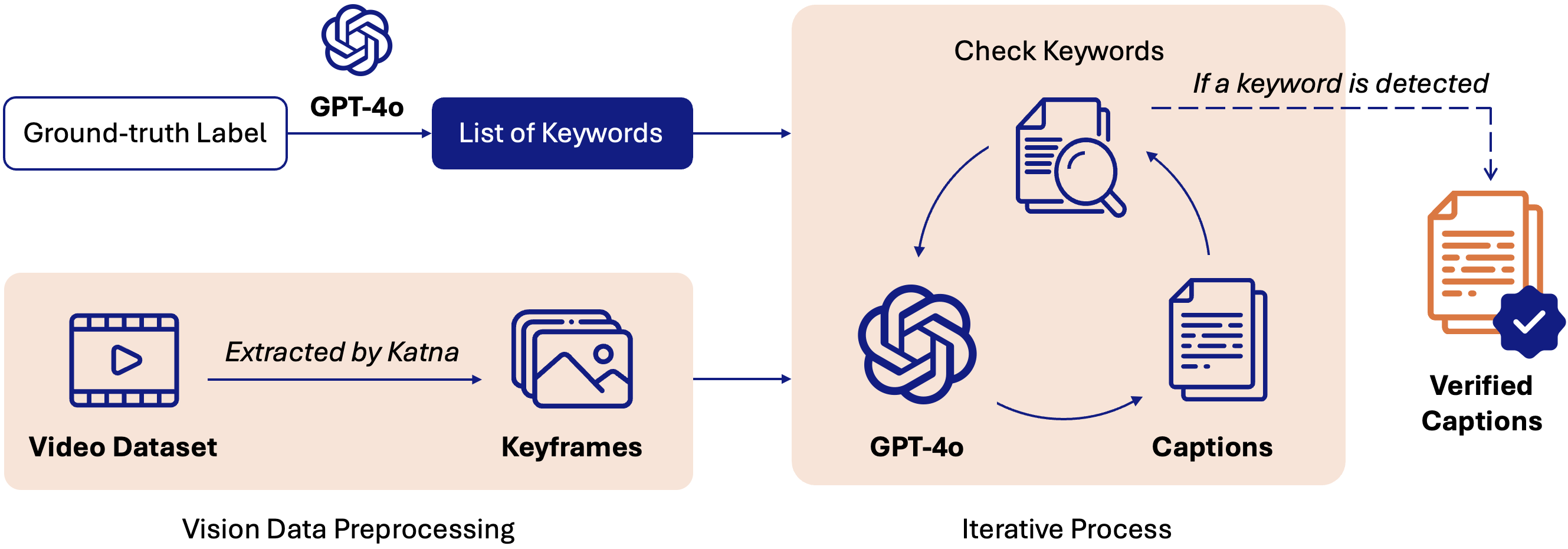}
    \caption{Overview of the descriptive caption generation framework.}
    \label{fig:enter-label}
\end{figure}

This study proposes a framework that integrates a VLM to generate descriptive captions from visual inputs and their corresponding ground-truth labels. The framework follows a systematic and iterative process to ensure that the generated captions align closely with the data set’s ground truth. The framework begins by preparing the following elements:

\begin{itemize}
    \item \textbf{Keyframes:} Keyframes are extracted from the video dataset using Katna~\cite{katna}, an automated library designed for efficient video and image processing tasks. The library focuses on keyframe extraction, video compression, and intelligent resizing and cropping. In its video module, Katna identifies keyframes by analyzing frame differences, brightness scores, and entropy, and by applying clustering techniques such as K-Means based on image histograms.

    \item \textbf{Keywords:} GPT-4o is utilized to create a curated list of keywords that are closely associated with the original ground-truth labels for each activity. For example, the curated list of keywords generated by GPT-4o for the ground-truth label \texttt{"Cook\_Cleanup"} is shown below:
\end{itemize}

\begin{lstlisting}[label={lst:code_example}, breaklines=true, breakatwhitespace=true]
    "Cook_Cleanup": ["cook", "organize", "wipe", "dish", "dishwasher", "load", "clean", "area", "counter", "table","stove", "tidy", "kitchen", "scrub", "wipe", "disinfect", "spotless", "neat"]
\end{lstlisting}

As illustrated in Figure~\ref{fig:enter-label}, the visual input, ground-truth labels, and associated keywords are provided to the VLM. GPT-4o is selected for this task due to its advanced capabilities in multimodal reasoning and articulate text generation. The model first generates an initial caption based on the given inputs. An algorithmic tool then checks whether the generated caption explicitly incorporates any of the associated keywords. If none are detected, the prompt is refined to prioritize the importance of integrating the keywords, and GPT-4o is re-invoked to produce a revised caption. This iterative process continues until a caption is generated that contains at least one relevant keyword, ensuring closer alignment with the ground-truth labels. Finally, an automated verification step is performed, confirming that 100\% of the generated captions include one or more of the corresponding keywords associated with each activity.

\section{Experiments and Analysis}

\subsection{Evaluation Methods}
The outputs of VLMs are not restricted to a predefined set of activity classes. Therefore, selecting an appropriate evaluation method is critical for HAR with VLMs. In this study, four distinct evaluation approaches are used to ensure a comprehensive assessment of the performance of VLM in the context of recognition of activities. Each method offers its own strengths and limitations. All approaches utilize the generated textual ground-truth captions as a benchmark to evaluate the model's outputs. 

\subsubsection{Keyword Matching} A predefined set of representative keywords is curated for each activity class in the dataset. The captions generated by the VLM under evaluation are analyzed to determine the presence of these keywords. Although this method provides a straightforward measure of activity detection, it may fail to recognize accurate descriptions that utilize alternative phrasing or vocabulary not included in the keyword list.

\subsubsection{VLM-as-Judge} An additional VLM is employed as an evaluator. This model receives visual input, the ground truth caption, and the caption generated by the VLM under evaluation. The evaluator assesses the validity of the generated caption in relation to the ground truth and is instructed to output only True or False. This approach has two notable limitations: it is computationally expensive, and it depends on the evaluator model's ability to accurately judge its performance, which may be compromised if the model lacks sufficient visual understanding. Furthermore, the evaluator is susceptible to judgment errors even when the ground truth is explicitly included in its prompt.

\subsubsection{BERTScore} The BERTScore~\cite{bertscore} metric is used to assess the semantic similarity between ground-truth captions and generated captions. It computes precision, recall, and F1 scores by comparing token-level embeddings between the two texts. In this study, only the precision metric is used. If the precision score between a ground-truth caption and a generated caption exceeds 0.9, it is considered correct recognition. Although BERTScore provides a nuanced evaluation of caption quality, it can sometimes yield misleading results, particularly when generated captions describing incorrect activities exhibit greater structural similarity to the ground truth than those describing the correct activity.

\subsubsection{Cosine Similarity} 
In this method, the \textit{all-MiniLM-L6-v2} embedding model~\cite{embedding_model4similarity} is used to encode both the ground-truth captions and the outputs generated by the VLMs. Subsequently, the cosine similarity between the two embedded vectors is calculated. The formula is given by:

\[
\text{cosine\_similarity}(A, B) = \frac{A \cdot B}{\|A\| \|B\|}
\]

where \(A\) and \(B\) are the embeddings (vector representations) of the ground-truth caption and the output of VLMs, respectively, and \( \| \cdot \| \) represents the Euclidean norm of the vector. 

A threshold of 0.5 was empirically determined based on the manual evaluation of 20 examples, where captions with similarity scores above the threshold were considered correct, and those below were regarded as incorrect.

\subsection{Experiment Settings}
This experiment evaluates three selected state-of-the-art open-source models: Llama3.2-Vision-11B~\cite{touvron2023llamaopenefficientfoundation}, DeepSeek-VL2-Small~\cite{deepseek-vl2}, and InternVL2.5-8B~\cite{chen2024internvlscalingvisionfoundation}. These models are compared against a larger proprietary model, GPT-4o. Given the importance of the number of analyzed frames in providing richer context for human activity recognition (HAR) tasks, while balancing the associated computational costs, the number of input frames was set to two for all models, except for Llama3.2-Vision-11B, which accepts only a single frame as input due to its architectural constraints.

The decision to prioritize open-source models is driven by the need to ensure compliance with data privacy and security standards, particularly critical when working with sensitive healthcare-related data. Nevertheless, GPT-4o is included as a benchmark due to its recognized leadership among VLMs and its relevance for comparative evaluation.

The evaluation of the Toyota Smart Home dataset is carried out in two phases. In the first phase, all models are tested on a subset consisting of 10 samples per class. The second phase follows the standard evaluation protocols of the dataset, involving cross-subject (CS) and cross-view (CV) evaluations. For the CS evaluation, data from 11 subjects (IDs: 3, 4, 6, 7, 9, 12, 13, 15, 17, 19, and 25) are used for training, while the remaining 7 subjects are reserved for testing. For the CV evaluation, the train-test split is determined based on the camera used for recording. Data from Camera 2 are used for testing. Two cases are considered for CV evaluation: CV1, where only Camera 1 recordings are used for training, and CV2, where recordings from Cameras 1, 3, 4, 6, and 7 are used for training. CV evaluation focuses exclusively on the 19 activities captured by both Camera 1 and Camera 2. Following the dataset protocol, the evaluation metric used in the second phase is Mean Class Accuracy (MCA), which computes the accuracy for each class individually and then averages these values to obtain the final score.

\subsection{Results and Discussion} 
This section presents the evaluation outcomes for the Toyota Smarthome Dataset. We analyze the effectiveness of different evaluation metrics, compare model performances, and discuss key observations regarding their strengths and limitations.

\subsubsection{Toyota Smarthome Dataset - Phase 1:} This phase primarily aims to assess the reliability of the evaluation methods. Only those methods validated through this phase will be selected for use in Phase 2. 

\begin{table}
    \caption{Measured accuracy across four evaluation methods on a subset of the Toyota Smarthome Dataset}
    \setlength{\tabcolsep}{6pt}
    \centering
    \begin{tabular}{lcccc}
        \hline
        Evaluation Methods & GPT-4o & InternVL-2.5 & DeepSeek-VL2 & Llama3.2-Vision \\
        \hline
        Keywords    & 55.0 & 62.4 &  56.7 & \textbf{67.4}  \\
        BERTScore & 100.0  & 100.0  &  100.0 & 100.0  \\
        Similarity  & \textbf{78.2}  & 73.2 &  67.1 & 54.0  \\
        VLM-as-Judge   & 13.0  & \textbf{53.0} &  32.0 & 24.5  \\
        \hline
    \end{tabular}
    \label{tab:model_comparison}
\end{table}

Table \ref{tab:model_comparison} indicates that BERTScore can be misleading, as it rated all model-generated outputs as correct. This likely occurs because BERTScore does not focus solely on the activity described in the sentence, but instead assigns equal importance to all tokens. For example, in the video \texttt{Usetelephone\_p02\_r00\_v15\allowbreak\_c06.mp4}, InternVL-2.5 generated the caption: \textit{"The person is looking at a cup on the counter in the kitchen."}, while the ground truth caption is: \textit{"The person is using the phone."} Although the generated caption does not accurately capture the main activity, BERTScore considered it correct and assigned a high precision score of \textit{0.9215} (where 1 indicates identical sentences). As a result, all samples were classified as True by BERTScore.

\begin{figure}
    \centering
    \includegraphics[width=0.95\columnwidth]{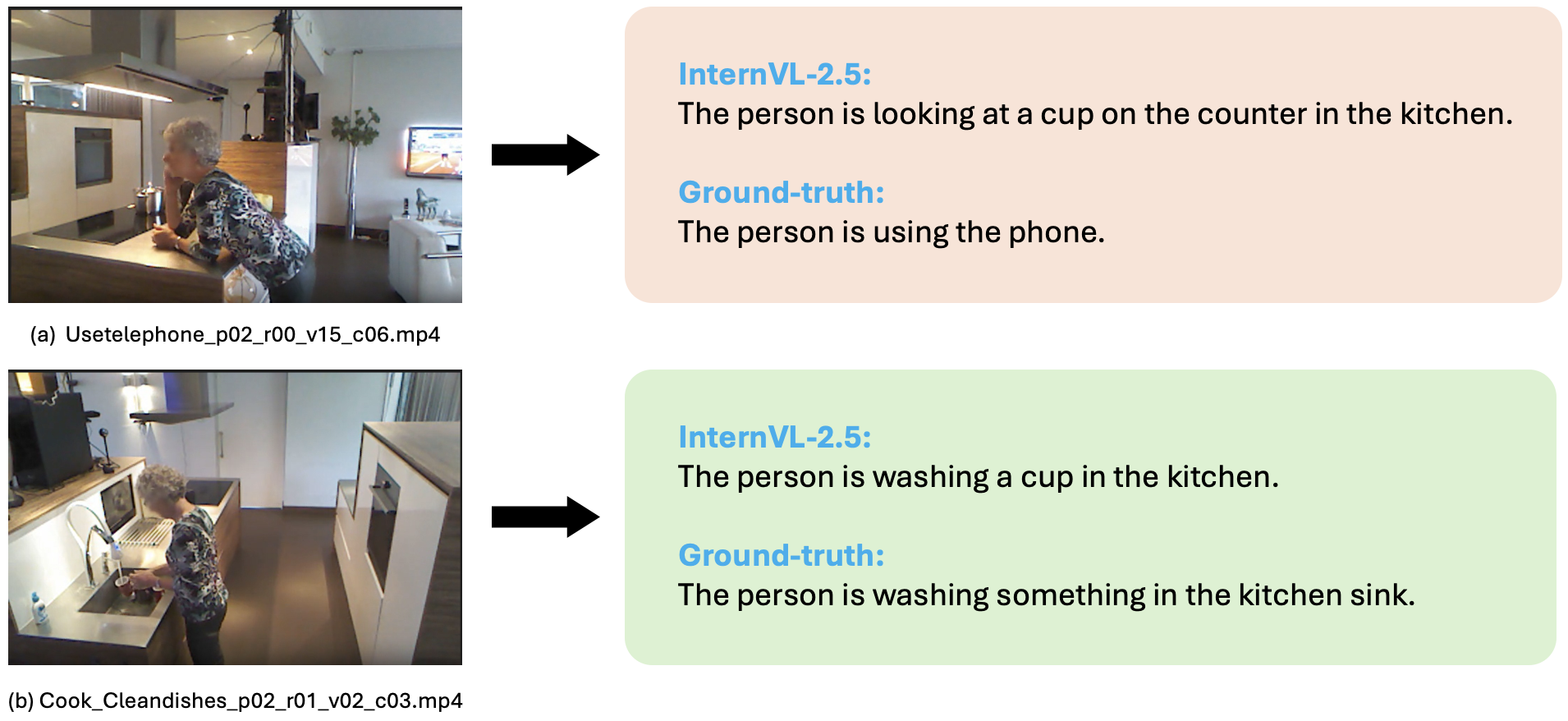}
    \caption{Comparison of InternVL-2.5 outputs and ground-truth captions}
    \label{fig:sample}
\end{figure}

The performance of the \textit{VLM-as-Judge} approach was lower than expected, even though it employed GPT-4o, the largest and most capable model among those evaluated. For instance, considering the video \texttt{Cook\_Cleandishes\_p02\_r01\allowbreak\_v02\_c03.mp4}, InternVL-2.5 generated the caption: \textit{"The person is washing a cup in the kitchen."} The correct (ground truth) caption is: \textit{"The person is washing something in the kitchen sink."} Although the generated caption accurately captures the core activity, the system judged it as incorrect. Nevertheless, this outcome offers a positive indication that the ground-truth caption dataset generated through the descriptive captioning framework is not biased toward GPT-4o’s outputs. The iterative refinement process within the framework played a critical role in mitigating potential biases during dataset construction, ensuring the correctness and fairness of the dataset.

The Cosine Similarity and Keyword Matching approaches appear to be the most reliable metrics. Therefore, these two are chosen to continue using in phase 2 of the experiment.

\subsubsection{Toyota Smarthome Dataset - Phase 2:} 
This phase adheres to the standard evaluation protocol of the dataset. Accordingly, the performance of VLMs is compared against that of traditional deep learning models for human activity recognition. Since deep learning models typically produce a single activity class as output, their evaluation results are directly aligned with the keyword matching and similarity methods described earlier. Following the study of Reilly et al.~\cite{pi}, seven vision-based deep learning models are selected for comparison with the VLMs. These models include AssembleNet++~\cite{assemble}, LTN~\cite{ltn}, VPN~\cite{vpn}, Video Swin~\cite{videoswin}, MotionFormer~\cite{motionformer}, TimeSformer~\cite{times}, and $\pi$-ViT~\cite{pi}.

\begin{table}
    \centering
    \caption{Mean class accuracy (MCA) on the Toyota Smarthome Dataset}
    \setlength{\tabcolsep}{6pt}
    \begin{tabular}{lcccc}
        \hline
        Models & Eval Methods & CS & CV1 & CV2 \\
        \hline
        \rowcolor{gray!20} \multicolumn{5}{c}{Deep Learning Models} \\
        AssembleNet++ \cite{assemble} & - & 63.6 & - & - \\
        LTN \cite{ltn} & - & 65.9 & - & 54.6 \\
        VPN++ \cite{vpn} & - & 69.0 & - & 54.9 \\
        Video Swin \cite{videoswin} & - & 69.8 & 36.6 & 48.6 \\
        MotionFormer \cite{motionformer} & - & 65.8 & 45.2 & 51.0 \\
        TimeSformer \cite{times} & - & 68.4 & 50.0 & 60.6 \\
        $\pi$-ViT \cite{pi} & - & 72.9 & 55.2 & \textbf{64.8} \\
        \hline
        \rowcolor{gray!20} \multicolumn{5}{c}{Vision Language Models} \\
        DeepSeek-VL2 & Keywords & 56.5 & 41.6 & 41.6 \\
        InternVL-2.5 & Keywords & 62.6 & 41.0 & 41.0\\
        Llama3.2-Vision & Keywords & 67.4 & \textbf{56.2} & 56.2\\
        \hline
        DeepSeek-VL2 & Similarity & 78.6 & 52.2 & 52.2 \\
        InternVL-2.5 & Similarity & \textbf{83.8} & \underline{56.1} & 56.1 \\
        Llama3.2-Vision & Similarity & 63.5 & 37.8 & 37.8\\
        \hline
    \end{tabular}
    \label{tab:experiment_results}
\end{table}

Table~\ref{tab:experiment_results} shows that, under the keyword matching evaluation method, all three open-source models, despite not being explicitly trained on the dataset and having access to only two keyframes per video for activity recognition, achieve competitive performance compared to traditional deep learning models. Remarkably, Llama3.2-Vision achieves a higher MCA (67.4\%) than several deep learning models, such as AssembleNet++, LTN, VPN++, and MotionFormer, in the CS evaluation. In the CV1 setting, Llama3.2-Vision also achieves the best performance (56.2\%) among all evaluated models. However, in the CV2 setting, after being trained with a larger amount of data, deep learning models outperform the VLMs, with the current state-of-the-art model, $\pi$-ViT, achieving an MCA of 64.8\%, thereby surpassing Llama3.2-Vision.

When evaluated using the cosine similarity method, which is considered a fairer approach for VLMs, all VLMs achieved higher MCA scores. In the CS evaluation, InternVL attained the highest MCA in the table at 83.8\%. DeepSeek-VL2 also achieved an MCA of 78.6\%, surpassing all deep learning models listed. In the CV1 setting, both InternVL2.5 and DeepSeek-VL2 continued to outperform traditional deep learning models. However, in the CV2 setting, they were surpassed by $\pi$-ViT, the current state-of-the-art model. Nevertheless, their performance remained higher than that of several other deep learning models.

A notable observation is that Llama3.2-Vision exhibits a significant drop in performance under the cosine similarity evaluation method. This decline can be attributed to its difficulty in controlling the length of its outputs. Despite being prompted to focus solely on the activities, Llama3.2-Vision often generates lengthy descriptions that attempt to capture every detail in the input frames. Under the keyword matching method, producing longer outputs increases the likelihood of including at least one curated keyword corresponding to the ground-truth activity label, which can lead to higher scores. However, in the cosine similarity evaluation, where only the semantic similarity between the generated caption and the ground-truth caption is considered, overly verbose outputs have a negative impact on performance. In contrast, DeepSeek-VL2 and InternVL2.5 consistently produce more concise and focused outputs, resulting in better alignment with the ground-truth captions.

\section{CONCLUSION}
This study proposes a descriptive caption generation framework to construct textual ground-truth captions for the Toyota Smarthome Dataset. The resulting human activity descriptive caption dataset is designed to enable a comprehensive evaluation of the human activity recognition (HAR) capabilities of Vision-Language Models (VLMs). Four evaluation methods were proposed, and experimental results verified that keyword matching and cosine similarity are the most reliable approaches.

In terms of VLM performance, despite not being trained on the dataset and having access to only two keyframes per video for activity recognition, VLMs, particularly InternVL2.5, achieved performance comparable to that of traditional deep learning models. These findings highlight the potential of VLMs to replace conventional deep learning models in HAR tasks, especially within assistive systems or Remote Health Monitoring Systems (RHMS), where consolidating multiple functionalities into a single model can significantly reduce computational demands. Furthermore, the experimental results suggest that the human activity descriptive caption dataset developed in this study can serve as a valuable resource for further fine-tuning VLMs and for providing a foundation for more rigorous and comprehensive evaluation of their performance in this domain.

\subsection*{Acknowledgements.} Abderrazek Abid and Thanh-Cong Ho contributed equally to this work.

\bibliographystyle{splncs04} 
\bibliography{main}
\end{document}